# AUTOMATIC TEXT SUMMARIZATION OF LEGAL CASES: A HYBRID APPROACH

Varun Pandya

*Department of Computer Science and Engineering, School of Technology,
Pandit Deendayal Petroleum University, India
varunpandya2004@gmail.com

## ABSTRACT

*Manual summarization of large bodies of text involves a lot of human effort and time, especially in the legal domain. Lawyers spend a lot of time preparing legal briefs of their clients' case files. Automatic Text summarization is a constantly evolving field of Natural Language Processing(NLP), which is a subdiscipline of the Artificial Intelligence Field.. In this paper a hybrid method for automatic text summarization of legal cases using k-means clustering technique and tf-idf(term frequency-inverse document frequency) word vectorizer is proposed. The summary generated by the proposed method is compared using ROGUE evaluation parameters with the case summary as prepared by the lawyer for appeal in court. Further, suggestions for improving the proposed method are also presented.*

## 1. INTRODUCTION

The amount of information and data has increased tremendously in every field over the years. With the limited human capacity of consuming information, it becomes paramount to extract only useful information out of the vast amount of unstructured data available. Thus, text summarization plays an important role in extracting useful pieces of information from possibly large bodies of text. Automatic Text Summarization is the process of producing textual summaries as output, given a certain textual document as an input, using the computational power of machine. It involves reducing a text document into a short set of words or paragraph that conveys the main meaning of the text[1]. The need for automatic text summarization in the field of Legal cases is continuously increasing with the number of cases piling up in the courtroom. Currently, a lot of effort goes into the manual drafting of case headnotes, synopses and summaries[2]. Legal systems across the world generate massive amounts of unstructured text every day; judges, lawyers, and case workers process and review millions of cases each year in the United States alone[3]. Preparing a legal case summary is a difficult task since the summary must contain relevant facts and precedents necessary to persuade the court in the favour of the plaintiff. Lawyers often spend many hours preparing legal briefs, which can be done by a computer in a few minutes. Thus, automatic text summarization is a necessity to overcome these challenges.

In this paper a novel hybrid approach of automatic text summarization in legal domain using clustering technique is proposed. The paper is organised as follows: Section 2 contains the overview and description of the terms associated with text summarization, Section 3 contains the overview of the existing approaches which are currently being used for text summarization, Section 4 contains the description of the proposed method, Section 5 consists of the information of data used for implementation and in Section 6 the approach is compared with the existing techniques of text summarization to get an insight in the performance of the proposed method. Section 7 contains the proposal for future work and Section 8 consists of the conclusion.

---

*Bachelor Student

## 2. BACKGROUND

Summarization is a challenging sub-task of the broader text-to-text generation field of natural language processing (NLP)[3].
Automatic text summarization is broadly of two types:

**1. Extractive Summarization:**
   This method involves selecting the most useful and relevant sentences from the document to provide a summary. Here the sentences originally present in the document are used to form the summary, with no new sentences being formed. Extractive summaries produce a set of the most significant sentences from a document, exactly as they appear[4].

**2. Abstractive Summarization:**
   In this type of summarization, new sentences are formed which are relevant to the document. Here the gist of the document is represented using computer generated sentences, having close resemblance to the summary as produced by humans. Abstractive text summarization captures the salient features of the text corpus and summary is generated by words/sentence that may/may not be in the input text corpus[5]. A lot of research is ongoing in this domain.

Depending upon the number of documents given as input, there are broadly two types:

**(a) Single Document Summarization:**
   Here a single document is given as an input to the text summarization model, and a summary of the document is produced as an output. The most useful and relevant sentences from the entire document are extracted and used to form a summary.

**(b) Multi Document Summarization:**
   Multi-document text summarization can be defined as a process of extracting relevant data and creating a short version of multiple source documents[6]. Here more than one document is given as input, and the sentences best representing the documents as a unit, are given as output.

The hybrid method proposed in this paper is based on extractive summarization of single document. A legal case is given as an input and a legal brief, best representing the useful and relevant facts, is produced as an output. The legal brief produced serves as an important tool for lawyers to present their case in the courtroom. They contain important information relevant to the case including all the facts and events which are necessary to win the case. It is as important to be economical to the time spent on the actual reading of the case as it is to be economical in the writing of the brief itself[7].

## 3. EXISTING APPROACHES

Different kinds of methods have been proposed for text summarization in the legal domain. The earliest of works in this area include the "Fast Legal EXpert CONsultant" (FLEXICON) system developed by Gelbart and Smith (Gelbartand Smith, 1991a). FLEXICON is keyword-based, referencing against a large database of terms to find important regions of text (Gelbart and Smith, 1991b). It served as a starting point for the text summarization approaches in legal domain. SALOMON, based on cosine similarity, proposed by Moens(Moens el al, 1999), served as an alternative to FLEXICON. Later the SUM and the Letsum(based on the thematic structure) projects served as an improvement on the previously proposed systems, to provide a structured representation of the legal briefs. Later a novel approach, LEXA, was proposed based on citation analysis. LEXA includes an interface for continued system learning using Ripple-down Rules (RDR), which allows domain experts to evaluate sentence selections live and agree or disagree with the selections[3].

A lot of systems for automatic text summarization have been proposed till date. These include CaseSummarizer, SUMMARIST, SMMRY and a lot of other proposed methods and algorithms. However, summarizing legal cases remains a challenging and daunting task because of the complexity of cases as well as the sheer size of legal cases. Many text summarization tools fail to produce summaries of large documents because of the complexity of algorithm used for summarizing the document. Thus, it is important to devise a method which not only gives optimal results but also gives them within a certain time limit. Also, it is important to specifically develop methods for text summarization of legal cases because of the size of legal cases, sometimes having more than 200 pages, which might result in system failure or an indefinite waiting period for complete execution. The system proposed in this paper is perfect for legal cases since it performs optimally and gives summary of required length, even when the case file has hundreds of pages.

## 4. THE PROPOSED SYSTEM

The implementation is carried out in three different steps: pre-processing, clustering of similar sentences, extracting top ranked sentences from each cluster

### 4.1. Pre-processing

Pre-processing of the document is a vital part of any text summarization approach. It is the foremost step of the proposed approach which includes removing of the unnecessary punctuations, merging the acronyms of a given sentences(example: F.C.A->FCA), removing extra periods(…..) which are often present in case files. Along with this the rich NLTK library of python has been used to remove stop words, tokenize the sentences and words and lemmatize them. After pre-processing, a clean document ready for clustering is produced.

### 4.2. Clustering

After pre-processing the case file, clustering of similar sentences is carried out. The clustering algorithm used in this approach is the k-means clustering algorithm. It is an iterative algorithm that executes many times with different *centroids* until it converges. Given $K > 0$ (where $K$ is the number of clusters) and a set of $N$ $d$-dimensional objects to be clustered:
- Clustering is the process of grouping a set of $N$ $d$-dimensional (2-dimensional, 3-dimensional, etc.) objects into $K$ clusters of similar objects.
- Objects should be like one another within the same cluster and dissimilar to those in other clusters.

In the proposed method, the rich sklearn library of python is used to vectorize the sentences. The tf-idf vectorizer of sklearn library is used to convert words to their vector representations. The tf-idf(term frequency-inverse document frequency) value of a word is given by:

$$w_{i,j} = tf_{i,j} \times \log\left(\frac{N}{df_i}\right)$$

$tf_{i,j}$ = number of occurrences of $i$ in $j$
$df_i$ = number of documents containing $i$
$N$ = total number of documents

Here clustering of sentences is performed at the document level meaning that similar sentences within the document will form clusters, such that intra-cluster distances are minimum and inter-cluster distances are maximum. The number of optimum clusters required for efficiency can be calculated using various algorithms like Elbow method, Information Criterion Approach, Silhouette width etc[8]. These clusters represent the similar types of sentences and hence certain number of sentences can be extracted from each cluster, depending upon their rank, such that the summary contains sentences obtained from different clusters, thus ensuring that the summary formed has less redundancy and contains sentences which can provide an overview of the entire legal case file. After the formation of clusters, the next step is to extract the top ranked sentences from each cluster.

### 4.3. Extraction of sentences from each cluster

For every sentence in each cluster, a rank is associated based on two features:

1. Tf-idf score of each sentence:
   The sum of tf-idf value of all words of the sentence is divided by the sum of tf-idf value for all the words present in the document.

   *sent_values = [sum(tf-idf of all words of sentence) / sum(tf-idf of all words of document)]*

2. Sentence similarity to title:
   The sentences similar to the title of the case generally present the important facts related to the case, thus aiding in generating a useful and relevant summary. The similarity score to the title is calculated as:

   *Title_score =( (length of similar words) * 0.1 ) / length of words in title)*

Here only the noun form words ['NN', 'NNS', 'NNP', 'NNPS' (Penn Tree Bank)] of the sentences and title are compared. The sentences belonging to a cluster are ranked based on the sum of these two scores (tf-idf and similarity of sentence to the title). The sentences having a higher score are ranked higher than the ones with lower score. Equal number of top ranked sentences are selected from each cluster for easier calculations. Thus, we get the optimal summary with the required number of sentences.

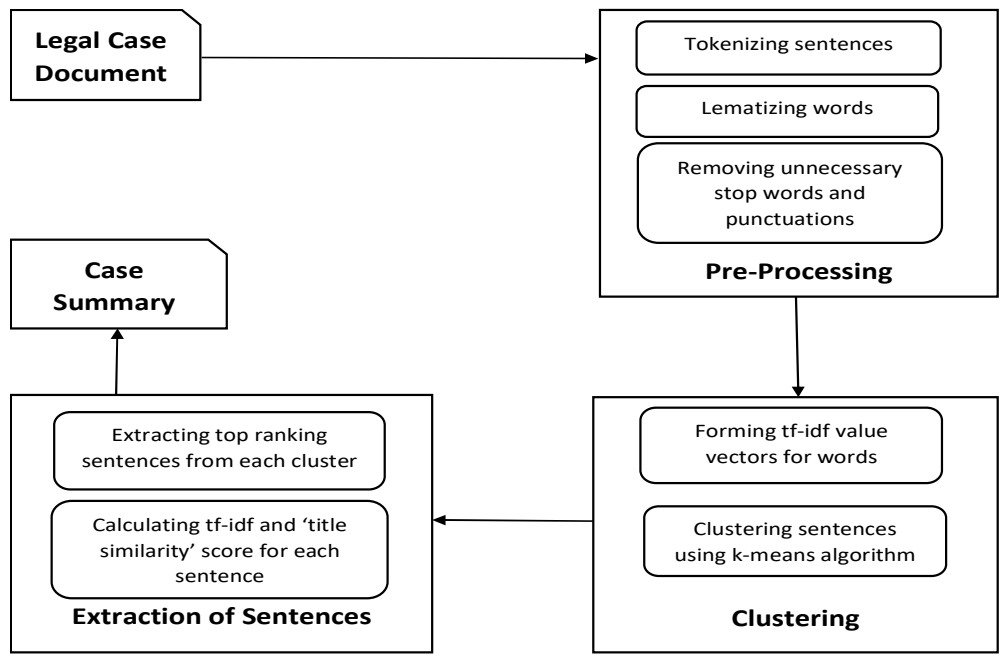

Figure 1. Framework of the proposed system

# 5. IMPLEMENTATION

For implementation, Auslii (http://austlii.edu.au) was used which consists of a database of Australian legal cases. The legal cases consist of the headnotes and summary within the case file. The legal cases downloaded in pdf format are given as input and the legal summary is produced as output by the system. The system was tested with a randomly selected legal case(Rush v Nationwide News Pty Ltd (No 7) [2019] FCA 496 (11 April 2019)) from the database. The case file consists of more than 200 pages, containing the facts and sequence of events required for convincing the jury. The summary provided with the case file consisted of 20 pages (approximately 150 sentences). This legal case was given as an input to the system and approximately 150 sentence summary was generated, in under 3 minutes, to match its length with the provided summary for better comparison.

# 6. EVALUATION

Summaries are difficult to evaluate because of their subjective nature. The relevance and usefulness of each sentence in the summary changes with the person evaluating them. Hence Lin et al. introduced a set of metrics called the ROUGE package in (Lin, 2004) that provide a pairwise comparison method for evaluating candidate summaries against human-provided ones[3]. ROUGE stands for Recall-Oriented Understudy for Gisting Evaluation and it counts the number of overlapping units such as n-gram, word sequences, and word pairs between the computer-generated summary to be evaluated and the ideal summaries created by humans [9]. The ROUGE metric has multiple variants(ROGUE N for N=1,2,3; ROGUE W; ROGUE L) and may be applied at the word, phrase, or sentence level. Here ROGUE 1, ROGUE 2, ROGUE L, ROGUE W are used as evaluation measures for comparing the proposed system with other existing systems. The Python Rogue library was used for it. Three automated tools were chosen for comparison TextSummarizer(http://textsummarization.net/text-summarizer), SplitBrain (https://www.splitbrain.org/services/ots), and ESummarizer(http://esummarizer.com/). Along with this the Text Rank algorithm was also used to summarize the legal case file, however it resulted in memory error due to computational complexity.

The ROGUE scores are as shown in the table 1

TABLE 1: ROGUE Scores Comparison

| ROGUE MEASURES | PROPOSED SYSTEM | | | TEXTSUMMARIZER | | | SPLITBRAIN | | | ESUMMERIZER | | |
|---|---|---|---|---|---|---|---|---|---|---|---|---|
| | PRECISION | RECALL | F-MEASURE | PRECISION | RECALL | F-MEASURE | PRECISION | RECALL | F-MEASURE | PRECISION | RECALL | F-MEASURE |
| ROGUE-1 | 27.62 | 28.16 | 27.88 | 12.2 | 9.71 | 10.81 | 21.51 | 19.42 | 20.41 | 21.78 | 21.36 | 21.57 |
| ROGUE-2 | 5.77 | 5.88 | 5.83 | 1.23 | 0.98 | 1.09 | 2.17 | 1.96 | 2.06 | 4 | 3.92 | 3.96 |
| ROGUE-L | 33.24 | 33.78 | 33.5 | 17.32 | 14.32 | 15.68 | 25.45 | 23.37 | 24.37 | 27.01 | 26.58 | 26.79 |
| ROGUE-W | 14.81 | 9.23 | 11.38 | 8.51 | 4.15 | 5.58 | 12.13 | 6.7 | 8.63 | 11.66 | 6.99 | 8.74 |

The proposed method in this paper performs favourably well against other existing methods. Since the proposed method is an unsupervised approach and is computationally favourably, it does not require training any data or creating corpus like it is required in some recently proposed approaches[10].
The reason why the proposed system performs favourably well is due to the extraction of sentences having the highest tf-idf scores and similarity to the title of the case, not just from the entire case file, but from each cluster of sentences obtained, thus ensuring least redundancy and similarity between the sentences generated in the final summary so that the final summary that gets generated contains sentences which help in representing the entire case file with least redundancy.

## 7. FUTURE WORK

The preliminary results of the proposed system serve as a promising means of providing legal summaries of case files. However, the proposed approach can be further improved by making the following addition:
- For determining the optimal number of clusters for k-means, certain algorithms can be added as an extension to the system for optimal results[8].
- For ranking the sentences after clustering, other than tf-idf and title similarity scores, parameters like position of sentence in the document and others can be used along with integration of other algorithms like Latent Dirichlet Algorithm or Graphical based approaches to generate sentences which have a higher overall rank.
- As an addition to the proposed system Thematic roles and Rhetorical Roles can be implemented to provide a structure to the document like it is required for legal case briefs.

## 8. CONCLUSIONS

The proposed method works favorably well against existing approaches. The summary generated closely resembles to the original summary as generated by the attorney of the case and can possibly be used in the court of law after further improvements. The summary generated by the proposed system after further revisions and structuring can possibly be used by attorneys for real time cases in the near future, after carrying out real world tests with attorneys which have not been conducted as of now. Since it is an unsupervised approach, consisting of clustering using k-means and extracting top ranked sentences from each cluster, and is computationally favorable, it provides a promising start towards developing a fully functional Legal Case Summarizer.

## 9. REFERENCES

[1]    Al-Hashemi, R.: "Text Summarization Extraction System (TSES) Using Extracted Keywords." International Arab Journal of e-Technology 1(4) (June 2010)

[2]    Marie-Francine Moens: "Summarizing court decisions" Information Processing and Management: an International Journal Volume 43 Issue 6, November, 2007   Pages:1748-1764

[3]    Polsley, Seth, Pooja Jhunjhunwala and Ruihong Huang. "CaseSummarizer: A System for Automated Summarization of Legal Texts." (COLING (2016)).

[4]    RafaelFerreiraa, Lucianode Souza Cabrala, Rafael DueireLins, Gabriel Pereira e Silvaa, Fred Freitas, George D.C. Cavalcantia, Rinaldo Lima, Steven J. Simske, Luciano Favaroc : "Assessing sentence scoring techniques for extractive text summarization" Expert Systems with Applications Volume 40, Issue 14, 15 October 2013, Pages 5755-5764

[5]    Nithin Raphal ; Hemanta Duwarah ; Philemon Daniel: "Survey on Abstractive Text Summarization" 2018 International Conference on Communication and Signal Processing (ICCSP)

[6]    Amol Tandel ; Brijesh Modi ; Priyasha Gupta ; Shreya Wagle ; Sujata Khedkar: "Multi-document text summarization - a survey" (2016 International Conference on Data Mining and Advanced Computing (SAPIENCE))

[7]    Casenote Legal Briefs for Torts, Keyed to Prosser, Wade Schwartz Kelly and Partlett

[8]    Trupti M. Kodinariya Dr. Prashant R. Makwana: "Review on determining number of Cluster in K-Means Clustering" (2013, IJARCSMS)

[9]    Lin, Chin-Yew. "ROUGE: a Package for Automatic Evaluation of Summaries." In Proceedings of the Workshop on Text Summarization Branches Out (WAS 2004), Barcelona, Spain, July 25 - 26, 2004.

[10]     Yogan, Jaya Kumar and Goh, Ong Sing and Halizah, Basiron and Ngo, Hea Choon and Puspalata, C Suppiah "A Review On Automatic Text Summarization Approaches." Journal Of Computer Science(2016), 12 (4). pp. 178-190. ISSN 1549-3636

[11]     Mieczyslaw A. Klopotek and Slawomir T. Wierzchon: Modern Algorithms of Cluster Analysis

[12]     Steven Bird, Ewan Klein and Edward Loper: Natural Language Processing with Python

[13]     Data Algorithms by Mahmoud Parsian

[14]     M. Saravanan • B. Ravindran : "Identification of Rhetorical Roles for Segmentation and Summarization of a Legal Judgment"(B. Artif Intell Law (2010) )

**Authors**

**VARUN PANDYA** is currently in third year, pursuing B.Tech in Computer Engineering from Pandit Deendayal Petroleum University.

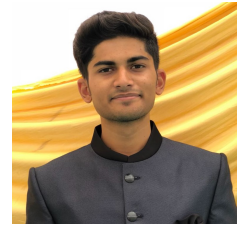